\documentclass[letterpaper, 10 pt, conference]{ieeeconf}

\IEEEoverridecommandlockouts

\usepackage{multicol}

\usepackage{amsmath}
\usepackage{amssymb}
\usepackage{booktabs}

\usepackage{bm}
\usepackage{booktabs}
\usepackage{multirow}

\usepackage{float}
\usepackage{caption} 
\usepackage{subfigure}

\usepackage{amsfonts}
\usepackage{graphicx}
\usepackage{textcomp}
\usepackage{todonotes}
\usepackage{csquotes}
\usepackage{xspace}
\usepackage{siunitx}

\usepackage{multicol}
\usepackage{multirow}

\usepackage{lipsum}
\let\labelindent\relax
\usepackage[shortlabels, inline]{enumitem}
\usepackage{wrapfig}

\usepackage{makecell}
\usepackage{hyperref}
\usepackage{algorithm}
\usepackage{algpseudocode}
\usepackage{svg}

\usepackage{cite}

\usepackage{pifont}
\usepackage{subcaption}

\title{\LARGE \bf
KineDepth: Utilizing Robot Kinematics for \\ Online Metric Depth Estimation
}

\author{Soofiyan Atar, Yuheng Zhi, Florian Richter, and Michael Yip%
    \thanks{All the authors are with the Electrical and Computer Engineering department, University of California San Diego.}%
    \thanks{This work was supported by the US Army Telemedicine and Advanced Technologies Research Center and NSF Career Award \#2045803.}%
}

\begin{document}

\maketitle
\thispagestyle{empty}
\pagestyle{empty}

\begin{abstract}
Depth perception is essential for a robot's spatial and geometric understanding of its environment, with many tasks traditionally relying on hardware-based depth sensors like RGB-D or stereo cameras. However, these sensors face practical limitations, including issues with transparent and reflective objects, high costs, calibration complexity, spatial and energy constraints, and increased failure rates in compound systems. While monocular depth estimation methods offer a cost-effective and simpler alternative, their adoption in robotics is limited due to their output of relative rather than metric depth, which is crucial for robotics applications.
In this paper, we propose a method that utilizes a single calibrated camera, enabling the robot to act as a ``measuring stick" to convert relative depth estimates into metric depth in real-time as tasks are performed. Our approach employs an LSTM-based metric depth regressor, trained online and refined through probabilistic filtering, to accurately restore the metric depth across the monocular depth map, particularly in areas proximal to the robot's motion.
Experiments with real robots demonstrate that our method significantly outperforms current state-of-the-art monocular metric depth estimation techniques, achieving a 22.1\% reduction in depth error and a 52\% increase in success rate for a downstream task.

\end{abstract}

\section{INTRODUCTION}
Robot manipulators are quickly becoming a commodity technology that can carry out laborious human tasks, such as household cleaning, food preparation and dishwashing, pantry stocking items, etc~\cite{SUOMALAINEN2022104224, ravichandar2020recent, jiang2022review}. The applications are broad, from home-care robot assistants\cite{home_assistant_sawik} to robots that operate and assist in the operating room\cite{operating_room}. In interacting with the world, the positional accuracy of the robot and its estimate of targets in the environment are basic necessities such that robot end-effectors can accurately and precisely reach, grasp, manipulate, and potentially deform objects and/or environments. 

Traditionally, 3D sensors such as RGB-D cameras and stereo-cameras were used to acquire geometrical and spatial information in the workspace and of objects of interest~\cite{wang2024visualroboticmanipulationdepthaware}. As a standard approach, robotic manipulation systems with depth sensors have been proven to work well in a wide variety of scenarios~\cite{shridhar2022perceiveractormultitasktransformerrobotic, bao2023dexartbenchmarkinggeneralizabledexterous, xu2023unidexgraspuniversalroboticdexterous, mahler2019learning, Li_2024}. 

However, there are benefits and limitations to 3D sensors. Technologies such as structured light, stereo vision, or LiDAR, can have readings thrown off by reflective surfaces, transparent objects, ambient light interference, and limited textural features~\cite{wasenmuller2016comparative}. They also tend to have more defined and limited operating ranges that have diminishing accuracy at both close and far distances~\cite{khoshelham2012accuracy}.

\begin{figure}[t]
    \centering
    \includegraphics[width=0.5\textwidth]{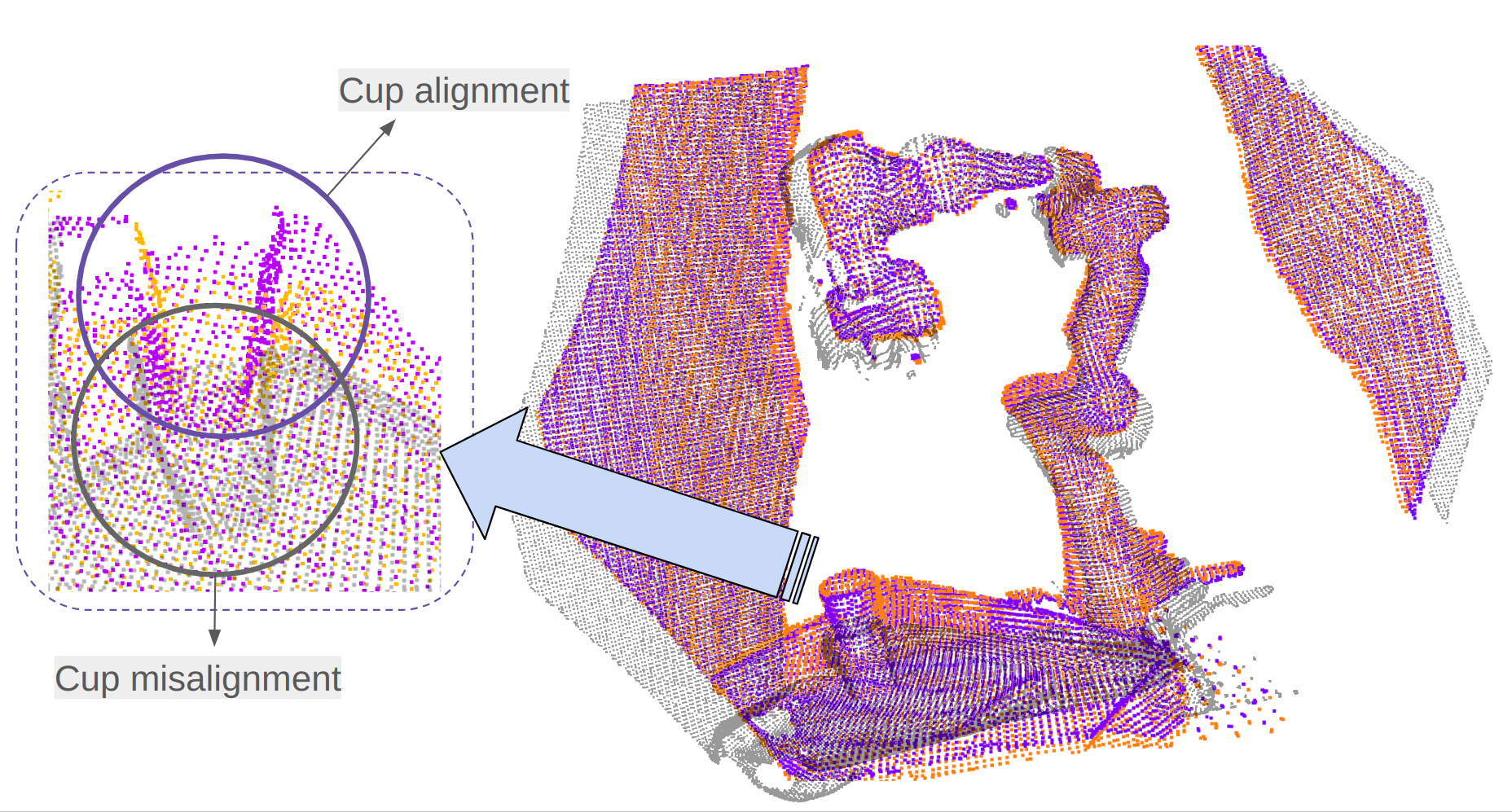}
    \caption{Visualization of point clouds for different depth estimation methods applied to a target object (cup). The orange point cloud represents KineDepth, the gray shows scaled relative depth from DepthAnything v2~\cite{yang2024depthv2}, and the purple represents ground truth metric depth. In the zoomed-in view, KineDepth closely aligns with the ground truth, while the relative depth shows significant local and global misalignment.}
    \label{fig: point cloud comparison}
\end{figure}

Monocular depth estimation (MDE) methods have more recently become available, where depth may be estimated from a single RGB image and a learned model~\cite{ming2021deep}. This alleviates some issues associated with depth sensors and makes monocular RGB with a learned depth model an interesting and more widely accessible alternative. In support of significant potential in MDE, very recently, training of foundation models with large-scale datasets has shown significant leaps in resolving depth estimates across a highly diverse range of environments with a single model~\cite{patni2024ecodeptheffectiveconditioningdiffusion, Wang_Liang_Xu_Jiao_Yu_2024}.

Despite the advances in MDEs, a main challenge of their use in robotic control applications is that MDEs typically yield relative depth estimates, which, while useful, suffer from some practical drawbacks, such as having no sense of the scale of a scene. While \textit{metric} monocular depth estimators (MMDEs) are very recently becoming available and provide absolute depth estimates~\cite{hu2024metric3dv2versatilemonocular, piccinelli2024unidepthuniversalmonocularmetric, bhat2023zoedepthzeroshottransfercombining, yin2023metric3dzeroshotmetric3d}, their estimates are not as stable when used live, can vary frame to frame under dynamic environments, and are still susceptible to inaccuracies due to noise in camera parameters. This inaccuracy creates real problems for robots with solving for the correct inverse kinematics, image Jacobians, and evaluating workspace reachability for driving a robot end effector to a target. 
Thus, this limits the ability of the robot to do manipulation tasks requiring higher accuracy and precision.

To this end, we propose a \textit{KineDepth}, a novel methodology that derives accurate metric depth from monocular RGB images by incorporating the reference geometry of a robotic manipulator. Consider your robot as a constant measuring stick in the scene, with its known geometry -- it can be used as a local reference for metric depth estimation.
Central to our approach is the integration of two aspects: (1) an off-the-shelf depth estimator such as Depth Anything V2~\cite{yang2024depthv2} that provides relative depth estimation using monocular cameras, without requiring fine-tuning, and (2) a novel \textit{depth scale regressor} --- a mathematical function that maps relative depth to metric depth that dynamically adapts based on the manipulator's kinematic chain as it approaches a target in the task space. 
As the manipulator moves closer to the target, the precision of the reference geometry increases, thereby enhancing the accuracy of metric depth estimation.

\begin{figure*}[t]
    \centering    \includegraphics[width=\textwidth]{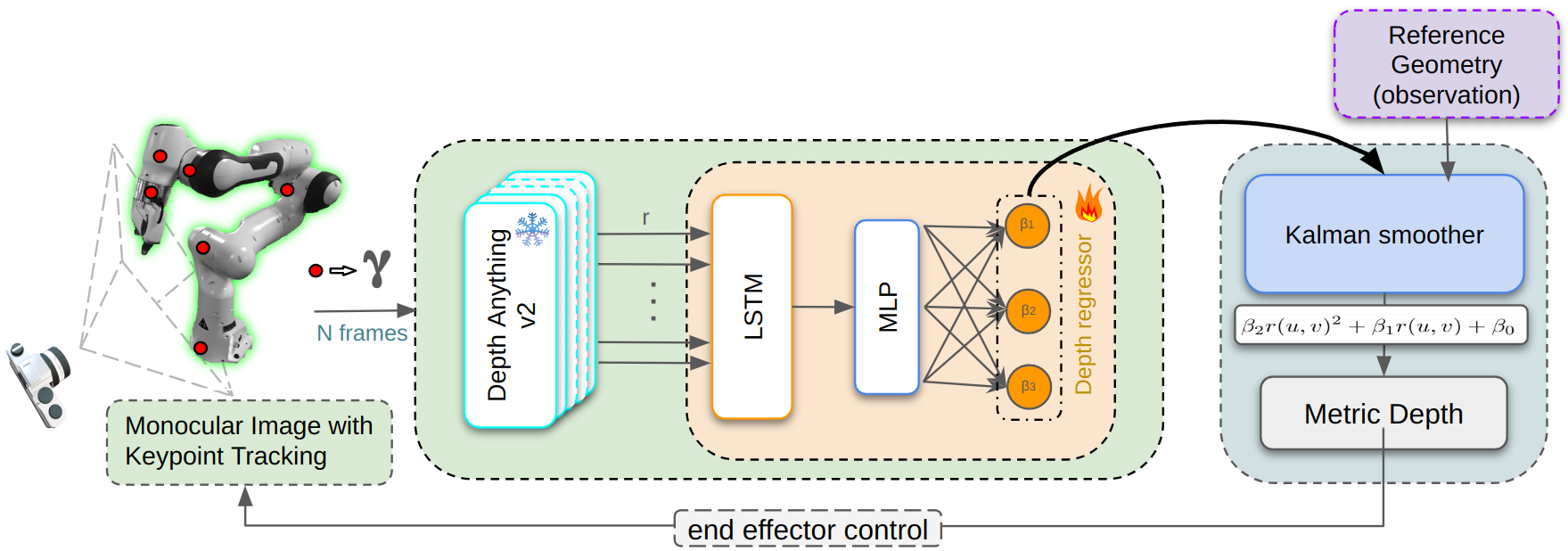}
    \caption{
    Our framework leverages the robot as a geometric reference to convert relative depth from monocular depth estimation techniques such as Depth Anything V2 to metric depth.
    The geometric references are keypoints on the robot, $\gamma$, which are applied to update our LSTM and Kalman Filters to estimate our depth regressor parameters, $\beta$. The depth regressor parameters create a polynomial fit to finally convert the relative depth to metric depth.
    The metric depth is used as feedback in downstream robotic control and manipulation tasks such as end-effector control.
    }
    \label{fig:pipeline}
\end{figure*}

 With the ultimate goal of working in novel, unstructured environments, we focus the KineDepth approach on working online and without pre-training. 
The key contributions of the paper are summarized as follows:
\begin{itemize}
    \item \textbf{Novel Monocular Metric Depth Estimation  Framework:} We present a new MMDE framework that synergizes relative depth from a foundation model with observed robotic manipulator geometry.
    \item \textbf{Dynamic Depth Scale Regressor:} We introduce a dynamic depth scale regressor that adapts to the manipulator's kinematic chain, enhancing depth precision as the end-effector approaches the target.
    \item \textbf{Online Tracking and Filtering:} We provide KineDepth with online estimation and filtering so that it may be deployed without pretraining and, therefore, for unstructured and unseen environments. It leverages an LSTM network for temporal modeling, which predicts the depth scale regressor refined by a Kalman filter for consistency and noise reduction.
\end{itemize}
Moreover, through extensive evaluations of random trajectories and pick-and-place tasks, we demonstrate that our approach outperforms state-of-the-art models in depth prediction accuracy, providing a scalable solution using only monocular cameras for high-precision manipulation tasks.

\vspace{-3mm}
\section{RELATED WORK}

Monocular depth estimation has been a pivotal area of research in computer vision, particularly for its applications in robotics and 3D reconstruction. Traditional methods for Single-Image Depth Estimation (SIDE) ~\cite{yang2024depthanythingunleashingpower, hu2024metric3dv2versatilemonocular, piccinelli2024unidepthuniversalmonocularmetric, bhat2023zoedepthzeroshottransfercombining, guizilini2023zeroshotscaleawaremonoculardepth, yin2023metric3dzeroshotmetric3d} can be categorized into metric depth regression and relative depth estimation~\cite{yang2024depthv2, ke2024repurposingdiffusionbasedimagegenerators, gui2024depthfmfastmonoculardepth, patni2024ecodeptheffectiveconditioningdiffusion, Wang_Liang_Xu_Jiao_Yu_2024}. Metric depth models often suffer from overfitting due to training on singular datasets, leading to poor generalization in unseen environments or varying depth ranges. Relative depth models, trained using scale-invariant losses on diverse datasets, tend to generalize better but lack the metric scale necessary for applications requiring absolute depth measurements. Recent efforts aim to recover metric information by combining monocular depth estimation with additional modules or by reformulating depth regression as a classification task.

Advancements in distribution learning have led to methods that treat depth estimation as a combined classification-regression problem, reasoning about depth value distributions across images. Approaches like AdaBins~\cite{Farooq_Bhat_2021}, LocalBins~\cite{bhat2022localbinsimprovingdepthestimation}, and PixelFormer~\cite{agarwal2022attentionattentioneverywheremonocular} have improved depth estimation by adaptively discretizing predicted depth ranges into bins, allowing for more precise depth predictions.

Monocular Metric Depth Estimation (MMDE) using neural networks was pioneered by transformer-based architectures~\cite{ranftl2021visiontransformersdenseprediction}. Despite significant progress on benchmarks like NYU Depth v2~\cite{silberman2012indoor} and KITTI~\cite{geiger2013vision}, MMDE models often struggle with zero-shot generalization due to domain shifts in appearance and geometry.

To enhance generalization across diverse domains, recent methods incorporate camera awareness by integrating external camera parameters or normalizing outputs based on intrinsic properties~\cite{guizilini2021sparse, yin2023learning, piccinelli2024unidepthuniversalmonocularmetric}. However, these approaches typically rely on noiseless camera intrinsics and predefined back-projection operations, limiting their applicability, especially in the absence of accurate camera information.

The use of large-scale data has propelled advancements in depth estimation. Models trained on extensive datasets~\cite{ranftl2020robustmonoculardepthestimation, yin2021learning, eftekhar2021omnidata} demonstrate improved generalization. Self-supervised learning approaches~\cite{he2020momentum, chen2020simple} and large-scale pretraining~\cite{radford2021learning, oquab2023dinov2} have further enhanced model robustness. These methods are prone to domain shift in new environments, and they may overfit to dataset-specific biases while lacking the fine-grained accuracy needed for precise depth estimation, issues we aim to address in this paper.

Despite the progress, there remains a gap in achieving accurate metric depth estimation from monocular RGB images without relying on additional sensors or prior knowledge. Existing methods either generalize poorly to new environments or lack the metric scale necessary for precise applications in robotic manipulation and 3D reconstruction. 

\section{METHODS}

This section outlines the pipeline for Monocular Metric Depth Estimation (MMDE) as shown in~\autoref{fig:pipeline}.
At the core of this method is updating a depth regressor in an online manner, which maps relative depth to metric depth using the visual measurements of the robot manipulator with known kinematics as known geometry.

\subsection{Depth Regressor}

The depth regressor, $Z(u, v) = f(r(u, v), \boldsymbol{\beta})$ for pixel location $[u,v]^\top \in \mathbb{R}^2$, in our system is a function, $f(\cdot)$, that converts the relative depth image, $r(
\cdot)$, obtained from a monocular depth estimation model, such as Depth Anything v2 model~\cite{yang2024depthv2}, into metric depth, $Z(\cdot)$.
The depth regressor is parameterized by $\beta$ which we estimate in an online fashion using the robot as a geometric reference.
We constrained the depth regressor, $f(\cdot)$, to simple models to ensure stability when applying the regressor to downstream control tasks and evaluated various functions, including power law, rational, Gaussian, logarithmic, linear, and polynomial regressors.
We ran an experiment on fitting Depth Anythingv2 predictions to ground truth using data collected from our experiments and SciPy's \textit{optimize.fit} function.
The polynomial regressor demonstrated the highest accuracy in fitting, as shown in \autoref{tab:curve_fit}.

Thus, our depth regressor is defined as
\begin{equation}
\label{eqn:depth_regressor}
    Z(u, v) = \beta_2 r(u, v)^2 + \beta_1 r(u, v) + \beta_0.
\end{equation}
where coefficients $\beta = [\beta_2, \beta_1, \beta_0]^\top$ are polynomial coefficients that will be estimated in an online fashion.

\vspace{-3mm}
\begin{table}[b]
\centering
\caption{Table shows the accuracy fit for different regressor to fit the relative depth to estimate the metric depth}
\label{tab:curve_fit}
\begin{tabular}{lc}
\toprule
\textbf{Regressor} & \textbf{Fit Percentage}\\
\midrule
Gaussian Model & 26.77\%\\
Logarithmic Model & 81.61\%\\
Power-law Model & 85.05\%\\
Rational Model & 85.09\%\\
Linear model & 88.42\%\\
Polynomial (2nd order) model & \textbf{95.30}\%\\
\bottomrule
\end{tabular}
\end{table}

\subsection{Online Estimation}

We estimate the depth regressor parameters, $\beta$, in an online fashion using the robot as a geometric reference.
The robot geometric references are defined as keypoints, $\gamma^i \in \mathbb{R}^2$,
which correspond to the following observed metric depth, $o^i$, from the robot kinematics
\begin{equation}
    \label{eq:observation_equation}
    o^i = \left[
    \mathbf{T}_c^b \prod_{i=1}^{n} \mathbf{T}_{i-1}^{i}(\theta_i) p^i \right]_{z}
\end{equation}
where $\mathbf{T}_c^b \in SE(3)$ is the camera to base transform, $\mathbf{T}_{i-1}^{i}(\theta_i) \in SE(3)$ is the $i$-th joint transform with joint angle $\theta_i$, and $p^i$ is the translational offset of the 3D keypoint.
We assume knowledge of the extrinsics, $\mathbf{T}_c^b$ \cite{lu2023markerless}, robot kinematics $\mathbf{T}_{i-1}^{i}(\theta_i)$, and keypoint offsets $p^i$ to compute geometric references, $o^i$, for our online estimation of the depth regressor parameters, $\beta$.
Note $o^i$ should correspond to the metric depth at the keypoint location, $Z(\gamma^i)$.
The keypoints, $\gamma^i_t$, will be detected using CoTracker~\cite{karaev2023cotrackerbettertrack} at every timestep $t$, however other keypoint estimation techniques could be used \cite{lu2022pose}.
We present three different approaches for the online estimation: Kalman Filter, LSTM, and a Hybrid approach.
All three approaches are outlined in \autoref{alg:main_algorithm}.

\subsubsection{Kalman Filter}

To utilize a Kalman Filter for online parameter estimation of the depth regressor parameters, $\beta$, we define initialization, motion, and observation models.
The parameters are initialized as a Gaussian with a linear scale, and the motion model is assumed to be mean-zero and Gaussian since we do not expect the parameters to change drastically between timesteps.
Mathematically, the initialization and motion models are defined as
\begin{align}
    \beta_{0} \sim \mathcal{N}\left( [0, 1, 0]^\top, \Sigma_{0} \right) && \beta_{t} \sim \mathcal{N}\left( \beta_{t-1},\Sigma_{t} \right)
\end{align}
where $\Sigma_{0}$ is the initial covariance and $\Sigma_{t}$ is the motion model covariance.

Finally, the observation model is based on our reference geometry observations, $o^i$ defined in \autoref{eq:observation_equation}, and modelled as 
\begin{equation}
    \label{eq:observation_model}
    o^i_t \sim \mathcal{N}\left( \mathbf{H}\beta_t, \Sigma_o \right)
\end{equation}
where $\mathbf{H} = [r(\gamma^i_t)^2, r(\gamma^i_t), 1]$ and $\Sigma_o$ is the observation model covariance.
With all the models  linear and Gaussian, the Kalman Filter can be employed to estimate $\beta$.

\subsubsection{LSTM}

Our secondary estimation approach uses an LSTM which is trained in an online fashion to estimate the depth regressor parameters, $\beta$.
The LSTM takes as input the relative depth at each keypoint and predicts hidden state, $\textbf{h}_t$:
\begin{equation}
    \label{eq:lstm}
    \textbf{h}_{t}, \textbf{c}_t = \text{LSTM}\left([r(\gamma^1_t), \dots, r(\gamma^M_t)]^\top), \textbf{h}_{t-1}, \textbf{c}_{t-1}\right)
\end{equation}
where $\textbf{c}_t$ is the copy state.
The LSTM is followed by a multilayer perceptron (MLP) to ultimately predict the depth regressor coefficients, $\beta_t$, and denoted as
\begin{equation}
    \label{eq:mlp_pred}
    \beta_t = \text{MLP}(\textbf{h}_t) \; \; .
\end{equation}
Similar to the Kalman Filter technique, we train the model using the depth observations, $o^i$, with the following loss
\begin{equation}
    \sum_{i=1}^M \mathcal{L}_h\left(f \left(r(\gamma_t^i), \beta_t \right), o^i_t \right) + \mathcal{L}_{aux}
\end{equation}
where $\mathcal{L}_h$ is the Huber loss with threshold $\delta$,
\begin{equation}
\label{eqn:huber_loss}
\mathcal{L}_h(y, \hat{y}) = 
\begin{cases} 
\frac{1}{2}(y - \hat{y})^2 & \text{for} \ |y - \hat{y}| \leq \delta \\
\delta \left( |y - \hat{y}| - \frac{1}{2}\delta \right) & \text{otherwise}
\end{cases}
\end{equation}
$f(\cdot)$ is the depth regressor evaluated at the keypoint location of the relative depth map, $r(\gamma^i_t)$, with parameters $\beta_t$ predicted by the LSTM and MLP and $\mathcal{L}_{aux}$ is an auxiliary loss to help with training.
The auxiliary loss is applied to auxiliary outputs of the model: unique polynomial coefficients for each keypoint, $\beta^i_t$, and the metric depth at every keypoint, $m^i_t$.
Since the metric depth should be inferred from the polynomial coefficients, an additional MLP is applied to the head to predict $m^i_t$.
The auxilary loss is defined as
\begin{equation}
    \mathcal{L}_{aux} = \alpha_1 \sum_{i}^M \mathcal{L}_h \left(f\left( r( \gamma^i_t), \beta^i_t \right), o^i_t \right)  + \alpha_2 \sum_{i}^M \mathcal{L}_h (m^i_t, o^i_t) 
\end{equation}
where $\alpha_1, \alpha_2$ are weightings.
Furthermore, we append the last $N$ relative depth values at each keypoint to the input of the LSTM, i.e. $r(\gamma^i_{t-N}), \dots, r(\gamma^i_{t})$, and run the LSTM $\tau$ steps for one time-step $t$.

\subsubsection{Hybrid}

Experimentally, we found the LSTM estimation approach provided high accuracy but low precision, i.e. large fluctuations in the depth regressor parameters.
These fluctuations would make downstream tasks such as robotic manipulation become unstable.
To address this issue and improve the consistency of the results, we incorporate a Kalman Filter to smooth the LSTM-generated depth regressor parameters, $\beta_t$.

First, we maintain the online training described in the previous LSTM subsection.
Secondly, we apply a Kalman Filter by considering the LSTM a learned motion model.
The initialization and motion models are then defined as
\vspace{-1mm}
\begin{equation}
    \beta_t \sim \mathcal{N}\left( \text{MLP}(\textbf{h}_t), \Sigma_t \right)
\end{equation}
where $\textbf{h}_t$ is the hidden state from \autoref{eq:lstm}, the MLP infers $\beta_t$ from the hidden state as shown in \autoref{eq:mlp_pred}, and $\Sigma_t$ is the motion model covariance.
Finally, we directly apply the Kalman Filter update step by using the observation model described in \autoref{eq:observation_model}.

\begin{algorithm}[t]
    \caption{Online Estimation of $\beta$}
    \label{alg:main_algorithm}
    \begin{algorithmic}[1]
        \If{Kalman Filter}
            \State $\beta_{0} \gets [0, 1, 0], \textbf{P}_{0} \gets \Sigma_0$ \Comment{Kalman Estimation}
        \EndIf
        \If{LSTM or Hybrid}
            \State $\text{LSTM}, \text{MLP} \gets \text{initializeModel()}$
            \State $\textbf{h}_0 \gets 0, \textbf{c}_0 \gets 0$
        \EndIf
        \For{$t = 1$ to $T$}
            \State $I_t \gets \text{getRGBImage}()$
            \State $\theta_t \gets$ \text{getJointAngles}()
            \State $r_t \gets \text{monocularDepth}(I_t)$ \Comment{DepthAnythingV2}
            \State $\gamma_t \gets \text{keypoint}(I_t)$ \Comment{CoTracker}
            \State $o_t \gets \text{observedMetricDepth}(\theta_t)$
            \If{Kalman Filter}
                \State $\beta_{t}, \textbf{P}_{t} \gets \text{predictStep}(\beta_{t-1}, \textbf{P}_{t-1}, \Sigma_t)$
            \Else \Comment{LSTM or Hybrid}
                \For{$k = 1$ to $\tau$}
                \State $\mathbf{h}_{t} \gets \text{LSTM}(r_t(\gamma_t), \mathbf{h}_{t-1}, \mathbf{c}_{t-1})$
                \State $\beta_t \gets \text{MLP}( \mathbf{h}_{t})$
                \State $\mathcal{L} \gets \text{computeLoss}(\beta_t, o_t)$
                \State $\text{LSTM}, \text{MLP} \gets \text{backPropogate}(\mathcal{L})$
            \EndFor
            \EndIf
            \If{Kalman Filter or Hybrid}
                \State $\beta_{t}, \textbf{P}_{t} \gets \text{updateStep}(\beta_{t}, \textbf{P}_{t}, o_t, \Sigma_o)$
            \EndIf
        \EndFor
    \end{algorithmic}
\end{algorithm}

\begin{figure}[t]
    \centering
    \includegraphics[width=0.35\textwidth]{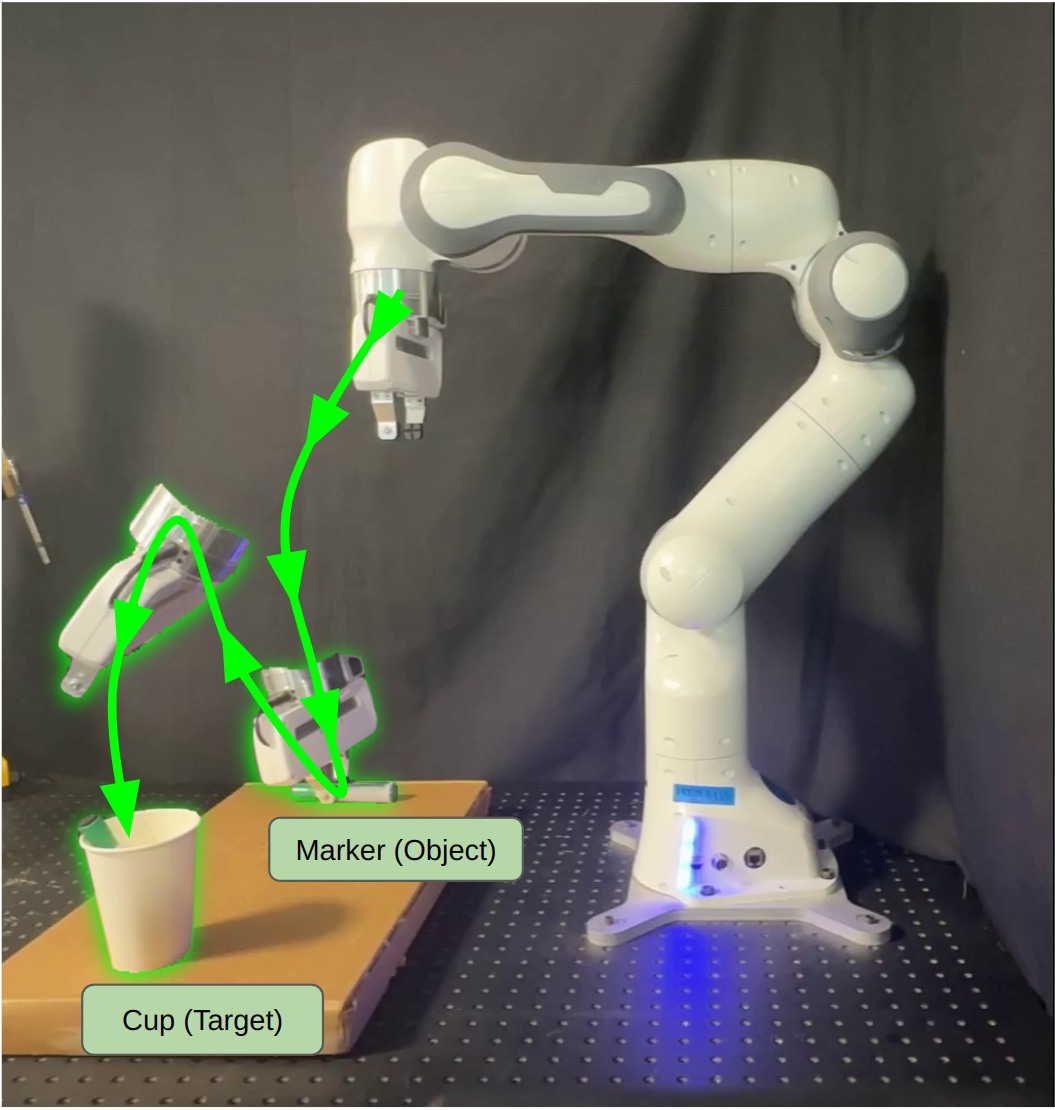}
    \caption{Pick-and-place task: Following a coarse trajectory using monocular depth estimation, where metric depth is calculated via KineDepth and tracked throughout the process.}
    \label{fig:pick_place_franka}
    \vspace{-5mm}
\end{figure}

\subsection{End Effector Control}

We employ the metric depth computed from \autoref{eqn:depth_regressor} using the estimated regression parameters, $\beta$, to control the end-effectors position in a closed-loop fashion.
We assume a goal in pixel space is given, $\mathbf{g}^p$, such as a cup detected in the image.
The goal is converted to 3D by inversing the pin-hole projection
\vspace{-3mm}
\begin{equation}
    \label{eq:compute_goal}
    \textbf{g}_t = f(r(\textbf{g}^p), \beta_t) \textbf{K}^{-1} \textbf{g}^p
\end{equation}
where $\textbf{K}$ is the camera intrinsics matrix.
While the end-effector could be computed using the robot kinematics, similar to the \autoref{eq:observation_equation}, we compute the 3D position of the end-effector using the depth regressor in the same manner as \autoref{eq:compute_goal} to minimize any potential bias from the estimated goal and end-effector which would lead the robot to miss the goal.
Therefore, the state is computed as
\begin{equation}
    \textbf{x}_t = f(r(\gamma^x_t), \beta_t)\textbf{K}^{-1} \gamma^x_t
\end{equation}
where $\gamma^x_t$ is the end-effector keypoint.
Finally, we define the action space as a change in end-effector, i.e., $\textbf{x}_t = \textbf{x}_{t-1} + \textbf{u}_t$.
A Linear Quadratic Regulator (LQR) is solved at every timestep with the following cost
\begin{equation}
    J = \sum_{i=t}^\infty \left( {(\mathbf{x}_i - \textbf{g}_t)}^\top \mathbf{Q} {(\mathbf{x}_i - \textbf{g}_t)} + \mathbf{u}_i^\top \mathbf{R} \mathbf{u}_i \right)
\end{equation}
where $\mathbf{Q}$ and $\mathbf{R}$ are weighting matrices.
Only one action is taken after solving the LQR, and then the entire process is repeated (compute goal, end-effector, and LQR) after estimating the latest depth regressor parameters, $\beta_t$.

\begin{figure*}[ht]
    \centering
    \includegraphics[width=0.9\linewidth]{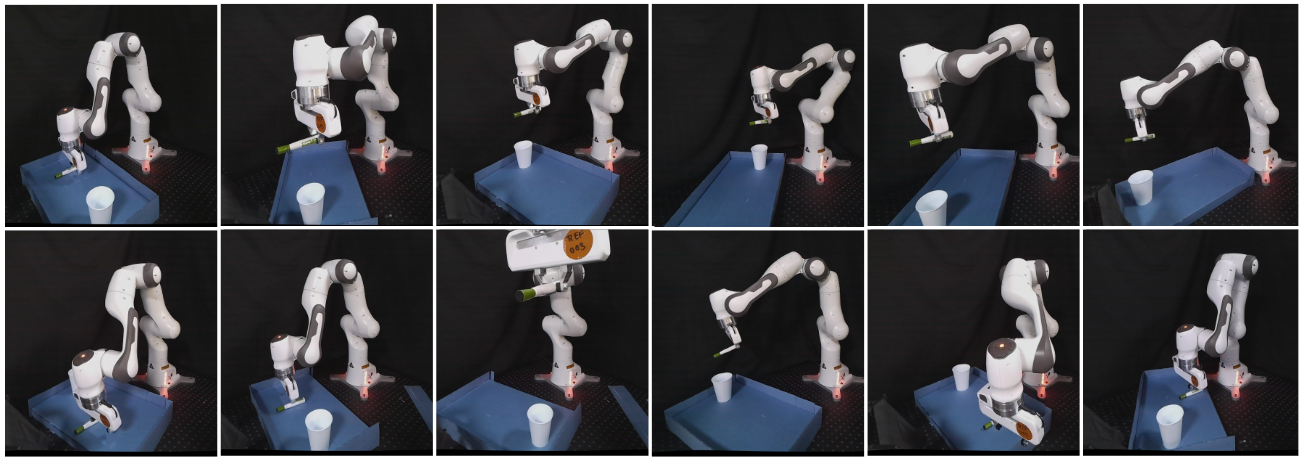}
    \caption{
    Physical Experiments: The upper row displays successful pick-and-place executions, while the bottom row illustrates failure cases. The figure covers a range of scene configurations and varying levels of grasping difficulty. Failures are primarily due to incorrect depth estimation or target poses prediction errors, leading to either missed grasps or inaccurate positioning of the marker.}
    \label{fig:results_success_failure}
    \vspace{-3mm}
\end{figure*}

    

\begin{figure*}[ht]
    \centering
    \subfigure[]{\label{fig:graph_methods_comp_a}\includegraphics[width=0.45\linewidth]{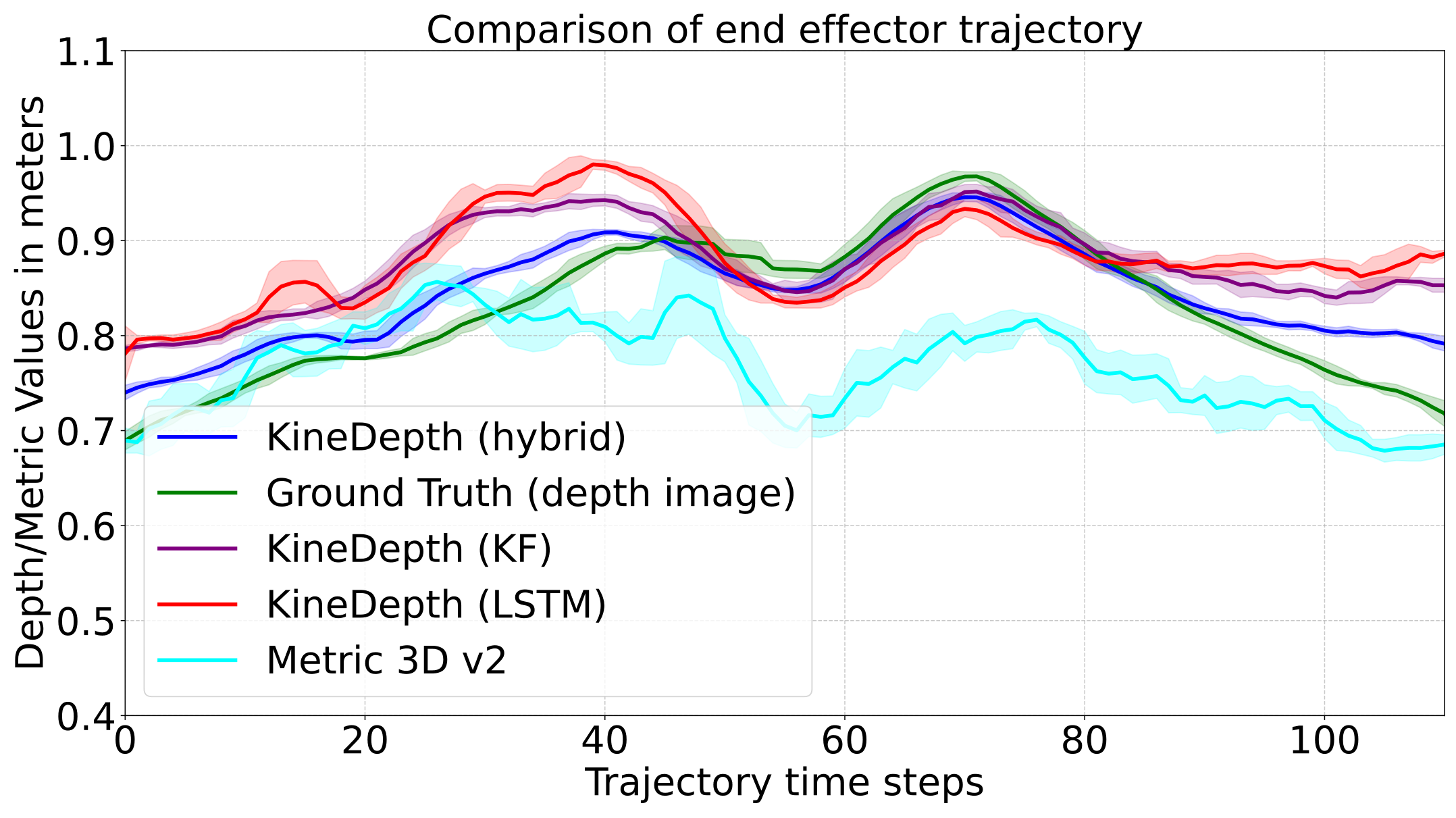}} \quad
    \subfigure[]{\label{fig:graph_methods_comp_b}\includegraphics[width=0.45\linewidth]{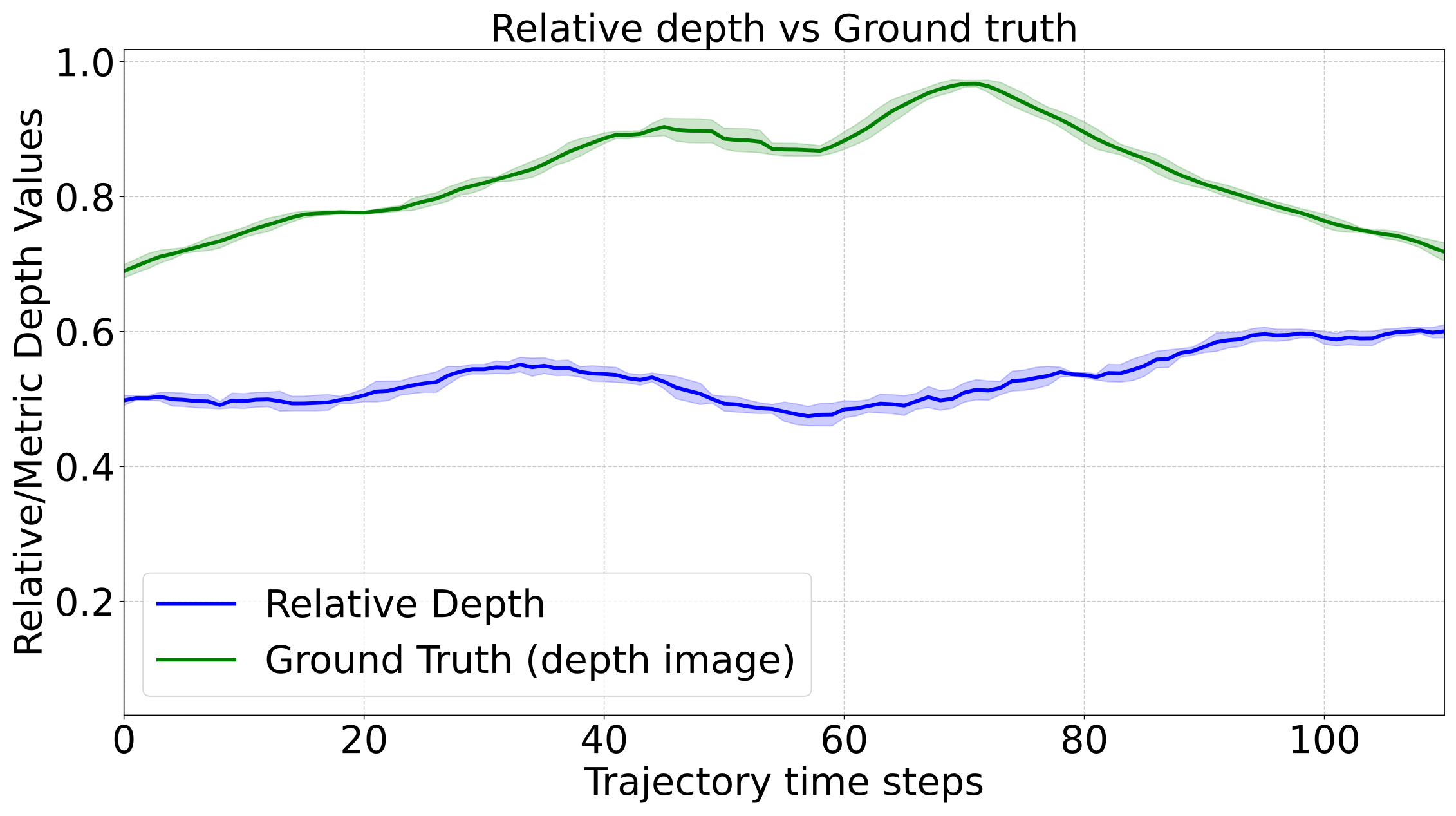}} \quad
    \vspace{-3mm}
    \caption{ (a) Trajectory comparison of the end effector across different methods, where the green curve represents the ground truth and the blue curve (KineDepth) shows the closest match. (b) Comparison of the relative depth trajectory with the ground truth, illustrating no correlation between the two.}
    
    \vspace{-3mm}
\end{figure*}

\vspace{-2mm}
\section{Experiments and Results}

\begin{table*}[ht]
\setlength\tabcolsep{0.3em}
\centering
\caption{Comparison of metric depth estimation accuracy across keypoints ($\gamma^1$ - $\gamma^5$) and overall scene error for different methods within the task space, including both the robotic manipulator and the task environment. The table reports errors for each method, with KineDepth variants (LSTM, KF, and hybrid) and traditional methods such as Metric 3D v2, Zoedepth, and DepthAnything v2.}
\label{tab:traj_comparison}
\begin{tabular}{l|c|c|c|c|c|c}
\toprule
\textbf{Method} & \textbf{$\gamma^1$ Error} & \textbf{$\gamma^2$ Error} & \textbf{$\gamma^3$ Error} & \textbf{$\gamma^4$ Error} & \textbf{$\gamma^5$ Error} & \textbf{Overall scene Error} \\
\midrule
DepthAnything v2~\cite{yang2024depthv2} & 1.544 $\pm$ 0.014 & 1.509 $\pm$ 0.035 & 1.608 $\pm$ 0.007 & 1.329 $\pm$ 0.148 & 1.326 $\pm$ 0.172 & 1.423 $\pm$ 0.059\\
UniDepth~\cite{piccinelli2024unidepthuniversalmonocularmetric} & 1.234 $\pm$ 0.017 & 1.255 $\pm$ 0.013 & 1.188 $\pm$ 0.010 & 1.214 $\pm$ 0.043 & 1.196 $\pm$ 0.044 & 1.242 $\pm$ 0.031\\
Zoedepth~\cite{bhat2023zoedepthzeroshottransfercombining} & 0.896 $\pm$ 0.004 & 0.706 $\pm$ 0.008 & 0.825 $\pm$ 0.008 & 0.885 $\pm$ 0.030 & 0.951 $\pm$ 0.028 & 0.878 $\pm$ 0.025\\
NeWCRF~\cite{yuan2022newcrfsneuralwindow} & 0.337 $\pm$ 0.002 & 0.515 $\pm$ 0.025 & 0.153 $\pm$ 0.004 & 0.450 $\pm$ 0.086 & 0.361 $\pm$ 0.078 & 0.195 $\pm$ 0.043\\
Metric 3D v2~\cite{hu2024metric3dv2versatilemonocular} & 0.150 $\pm$ 0.004 & 0.079 $\pm$ 0.003 & 0.044 $\pm$ 0.001 & 0.146 $\pm$ 0.046 & 0.074 $\pm$ 0.009 & 0.068 $\pm$ 0.003\\
\midrule
KineDepth (LSTM) & 0.121 $\pm$ 0.004 & 0.089 $\pm$ 0.002 & 0.059 $\pm$ 0.001 & 0.048 $\pm$ 0.001 & 0.035 $\pm$ 0.003 & 0.099 $\pm$ 0.005\\
KineDepth (KF) & 0.103 $\pm$ 0.000 & 0.073 $\pm$ 0.002 & 0.044 $\pm$ 0.000 & 0.060 $\pm$ 0.006 & 0.029 $\pm$ 0.001 & 0.061 $\pm$ 0.001\\
KineDepth (hybrid) & \textbf{0.035 $\pm$ 0.002} & \textbf{0.032 $\pm$ 0.001} & \textbf{0.036 $\pm$ 0.000} & \textbf{0.050 $\pm$ 0.002} & \textbf{0.011 $\pm$ 0.000} & \textbf{0.053 $\pm$ 0.001}\\
\bottomrule
\end{tabular}
\end{table*}

\begin{table*}[ht]
\setlength\tabcolsep{0.3em}
\centering
\caption{Comparison of metric depth estimation performance for different methods on a marker pick-and-place task involving various scene configurations: easy, intermediate, and hard. The table presents accuracy percentages for each configuration, as well as overall accuracy for each method.}
\label{tab:pick_place}
\begin{tabular}{l|c|c|c|c}
\toprule
\textbf{Method} & \textbf{Easy configuration} & \textbf{Intermediate configuration} & \textbf{Hard Configuration} & \textbf{Overall accuracy} \\
\midrule
ZoeDepth~\cite{bhat2023zoedepthzeroshottransfercombining} & 0.0\%& 0.0\%& 0.0\%& 0.0\%\\
UniDepth~\cite{piccinelli2024unidepthuniversalmonocularmetric} &0.0\% &0.0\% & 0.0\%& 0.0\%\\
DepthAnything v2~\cite{yang2024depthv2} & 0.0\%& 0.0\%& 0.0\%& 0.0\%\\
NeWCRF~\cite{yuan2022newcrfsneuralwindow} & 0.0\%& 0.0\%& 0.0\%& 0.0\%\\
Metric 3D v2~\cite{hu2024metric3dv2versatilemonocular} & 33.3\% & 25.0\% & 25.0\% & 28.0\%\\
\midrule
KineDepth (LSTM) & 33.3\% & 37.5\% & 37.5\% & 36.0\%\\
KineDepth (KF) & 81.8\% & 42.8\% & \textbf{66.7\%} & 64.0\%\\
KineDepth (hybrid) & \textbf{100\%} & \textbf{75\%} & 62.5\% & \textbf{80.0\%}\\
\bottomrule
\end{tabular}
\end{table*}

\vspace{-2mm}
\subsection{Implementation details}

Our experiment focuses on running KineDepth on Franka Robot, and all the experiments are compared using the same robot. \autoref{fig:pick_place_franka} depicts the setup of our environment where the camera used is Azure Kinect and the GPU used is Nvidia GTX 1080, which runs the DepthAnything V2 model~\cite{yang2024depthv2} with the Depth-Anything-V2-Large, CoTracker~\cite{karaev2023cotrackerbettertrack} and online training of LSTM network. The Azure Kinect is used to get the ground truth using the camera, and for the experiments, only the RGB channel was utilized. For the marker pick and place task, we used the setup depicted in \autoref{fig:pick_place_franka}, where the marker and cup were randomly placed.

In the LSTM implementation, we configured the key hyperparameters as follows. The learning rate was set to $1 \times 10^{-2}$, with a batch size equal to the number of visible keypoints $M$ in the current frame. The hidden state size was fixed at 128, and the network was trained for $\tau$ 20 time steps per update. $\Sigma_0$ is a identity matrix, $\Sigma_t$ are a diagonal matrix with 0.5 and $\Sigma_o$ are a diagonal matrix with 0.03. In Huber loss, the threshold is set at $\delta = 1.0$, while $\alpha_1$ and $\alpha_2$ are set to 1.0. Meanwhile, $N$ is set to 10 to store the trajectory history for the input of the LSTM. 

\vspace{-2mm}
\subsection{Experimental Setup}

The setup chosen here is to showcase the accuracy required to grasp the marker and place it in the cup where the goal given to the model was the pixel coordinates and intrinsic parameters. In our experiments, we compared our method against 3 variants of KineDepth, metric 3D v2~\cite{hu2024metric3dv2versatilemonocular}, Unidepth~\cite{piccinelli2024unidepthuniversalmonocularmetric}, NewCRFs~\cite{yuan2022newcrfsneuralwindow}, ZoeDepth~\cite{bhat2023zoedepthzeroshottransfercombining} and DepthAnything v2~\cite{yang2024depthv2} whose decoder is fine-tuned using ZoeDepth architecture. All the experiments are performed on real-world evaluation using the setup explained above.

\subsection{Trajectory Following for MMDE}

To evaluate the performance and robustness of metric depth estimation methods, we ran 25 random trajectories for each method and calculated the error at each keypoint, using the depth image as ground truth at the pixel coordinates $\gamma^i$. As shown in \autoref{tab:traj_comparison}, KineDepth achieved the lowest mean error across all keypoints and the overall scene, which includes the robotic manipulator and task space region. KineDepth (hybrid) had a 5.3 cm overall error, 1.5 cm lower than Metric 3D v2, which outperformed other MMDE methods. KineDepth (KF) estimates depth by modeling the depth regressor as the state in a Kalman filter, while KineDepth (LSTM) uses the predicted depth regressor directly from the network. Other methods had errors exceeding 19.5 cm for the overall scene.

\vspace{-2mm}
\subsection{Marker pick place task}

To quantitatively assess the methods, we designed a pick-and-place task with a marker and cup, requiring high precision. As shown in \autoref{fig:pick_place_franka}, the setup involves random marker and cup placements, with challenging configurations outlined in \autoref{tab:pick_place}. Misperceptions often occur when the marker is placed near the camera or diagonally, complicating trajectory planning. \autoref{fig:results_success_failure} shows different configurations, while \autoref{tab:pick_place} demonstrates KineDepth's superiority, with the hybrid version achieving 80\% accuracy on 25 pick-place tasks. On harder configurations, KineDepth (KF) slightly outperformed the hybrid version. Other baselines, apart from Metric 3D v2, had no success due to poor metric depth estimates, as supported by results compared in \autoref{tab:traj_comparison}.

\vspace{-2mm}
\section{Discussion and Conclusion}
As shown in \autoref{tab:traj_comparison} and \autoref{tab:pick_place}, KineDepth outperforms all baselines, with the hybrid version achieving even better results. Fig. \autoref{fig:graph_methods_comp_a} illustrates the end effector trajectory for various methods, where KineDepth closely matches the ground truth (depth camera). Other KineDepth variants also exhibit low error, while Metric 3D v2 produces a trajectory that significantly deviates from the ground truth. Furthermore, Fig. \autoref{fig:graph_methods_comp_b} clearly shows that the relative depth has no correlation with the true depth, indicating that simply scaling the relative depth does not achieve accurate metric depth estimation.

In conclusion, the proposed KineDepth framework introduces a depth regressor that converts relative depth $r$ to the metric depth with high accuracy in the task space region. Further, using Kalman filtering to smooth out the depth regressor predicted from the LSTM network boosted our performance in getting a more accurate estimation. KineDepth delivers improved metric depth accuracy in the task space region, and it achieved 1.5cm less error for the overall task space region as compared to Metric 3D v2. KineDepth (hybrid) provides 22.1\% higher accuracy than the state-of-the-art MMDE, achieving the lowest overall scene error of 0.053 m across all key points.

\bibliographystyle{IEEEtran}
\bibliography{root}

\end{document}